%% file: main.tex
\documentclass[10pt,twocolumn,letterpaper]{article}

\usepackage[pagenumbers]{cvpr} 

\usepackage{graphicx}
\usepackage{amsmath}
\usepackage{amssymb}
\usepackage{booktabs}
\usepackage{makecell}

\usepackage[pagebackref,breaklinks,colorlinks]{hyperref}

\usepackage[capitalize]{cleveref}
\crefname{section}{Sec.}{Secs.}
\Crefname{section}{Section}{Sections}
\Crefname{table}{Table}{Tables}
\crefname{table}{Tab.}{Tabs.}

\begin{document}


\title{Minimizing Maximum Model Discrepancy for Transferable  \\ Black-box Targeted Attacks}

\author{Anqi Zhao \space\space\space\space Tong Chu \space\space\space\space Yahao Liu  \space\space\space\space Wen Li\space\space\space\space Jingjing Li\space\space\space\space  Lixin Duan \space\space\space\space
\\
University of Electronic Science and Technology of China\\
{\tt\small \{zhaoanqiii, uestcchutong, lyhaolive, liwenbnu, lxduan\}@gmail.com,}
{\tt\small lijin117@yeah.net}
}

\maketitle

\begin{abstract}

In this work, we study the black-box targeted attack problem from the model discrepancy perspective. On the theoretical side, we present a generalization error bound for black-box targeted attacks, which gives a rigorous theoretical analysis for guaranteeing the success of the attack. We reveal that the attack error on a target model mainly depends on empirical attack error on the substitute model and the maximum model discrepancy among substitute models. On the algorithmic side, we derive a new algorithm for black-box targeted attack based on our theoretical analysis, in which we additionally \textbf{m}inimize the \textbf{m}aximum \textbf{m}odel \textbf{d}iscrepancy~(M3D) of the substitute models when training the generator to generate adversarial examples. In this way, our model is capable of crafting highly transferable adversarial examples that are robust to the model variation, thus improving the success rate for attacking the black-box model. We conduct extensive experiments on the ImageNet dataset with different classification models, and our proposed approach outperforms existing state-of-the-art methods by a significant margin. Our codes will be released. 

\end{abstract}

\section{Introduction}
\label{sec:intro}

Recently, researchers have shown that Deep Neural Networks (DNNs) are highly vulnerable to adversarial examples~\cite{szegedy2013intriguing,goodfellow2014explaining,papernot2016transferability}. It has been demonstrated that by adding small and human-imperceptible perturbations, images can be easily misclassified by deep-learning models. Even worse, adversarial examples are shown may have \emph{transferability}, \ie, adversarial examples generated by one model can successfully attack another model with a high probability~\cite{tramer2017space,liu2016delving,papernot2016transferability}. Consequently, there is an increasing interest in developing new techniques to attack an unseen black-box model by constructing adversarial examples on a substitute model, which is also known as \emph{black-box} attack~\cite{dong2018boosting,dong2019evading,xie2019improving,wu2020skip,inkawhich2019feature,Inkawhich2020Transferable,inkawhich2020perturbing,li2020towards}. 

While almost all existing black-box attack works implicitly assume the transferability of adversarial examples, the theoretical analysis of the transferability is still absent. To this end, in this work, we aim to answer the question of to what extent the adversarial examples generated on one known model can be used to successfully attack another unseen model. In particular, we are specifically interested in the targeted attack task, \ie, constructing adversarial examples that can mislead the unseen black-box model by outputting a highly dangerous specified class. We first present a generalization error bound for black-box targeted attacks from the model discrepancy perspective, in which we reveal that the attack error on a target model depends on the \emph{attack error on a substitute model} and \emph{the model discrepancy between the substitute model and the black-box model}. Furthermore, the latter term can be bounded by the maximum model discrepancy on the underlying hypothesis set, which is irrelevant to the unseen target model, making it possible to construct adversarial examples by directly minimizing this term and thus the generalization error. 

Based on the generalization error bound, we then design a novel method called Minimizing Maximum Model Discrepancy (M3D) attack to produce highly transferable perturbations for black-box targeted attack. Specifically, we exploit two substitute models which are expected to maintain their model discrepancy as large as possible. At the same time, we train a generator that takes an image as input and generates an adversarial example to attack these two substitute models and simultaneously minimize the discrepancy between the two substitute models. In other words, the generator and the two substitute models are trained in an adversarial manner to play a min-max game in terms of the model discrepancy. In this way, the generator is expected to generate adversarial examples that are robust to the variation of the substitute models, thus being capable of attacking the black-box target model successfully with a high chance.

We conduct extensive experiments on the ImageNet dataset using different benchmark models, where our M3D approach outperforms state-of-the-art methods by a significant margin on a wide range of attack settings. Especially, we show impressive improvements in the situations when the black-box model has a large model discrepancy from the substitute model, such as attacking the ResNet~\cite{he2016deep} model by crafting adversarial examples on a VGG~\cite{simonyan2015very} model. The main contributions of this paper are as follows:

\begin{itemize}
    \item We present a generalization error bound for black-box targeted attack based on the model discrepancy perspective. 
    \item We design a novel generative approach called Minimizing Maximum Model Discrepancy~(M3D) attack to craft adversarial examples with high transferability based on the generalization error bound.
    \item We demonstrate the effectiveness of our method by strong empirical results, where our approach outperforms the state-of-art methods by a significant margin.
\end{itemize}

\section{Related Work}
\label{sec:Related Work}

\noindent \textbf{Adversarial Attack:}
Many works on the adversarial attack have been proposed, since they reveal that deep neural networks (DNNs) are highly vulnerable to adversarial examples~\cite{dong2018boosting, xie2019improving, dong2019evading,wu2020skip,inkawhich2019feature,Inkawhich2020Transferable,poursaeed2018generative,carlini2017towards}. Generally, methods for adversarial attack can be divided into two branches: iterative perturbation methods and generative perturbation methods.

\emph{Iterative  Perturbations:} Iterative instance-specific attacks~\cite{dong2018boosting, xie2019improving, dong2019evading,wu2020skip,Wang_2021_ICCV, zhu2021rethinking} perturb a given sample by iteratively using gradient information. 
Though good performances have been achieved in white-box attack scenarios~\cite{goodfellow2014explaining,kurakin2018adversarial,madry2018towards}, their black-box transferability  are limited. Therefore, many methods have been proposed to improve the transferability against black-box models. Some works are inspired by model training process such as introducing momentum items to stabilize optimization ~\cite{dong2018boosting} or adapting Nesterov accelerated gradient into the iterative attacks ~\cite{lin2019nesterov}. Data augmentation methods have also been shown effective in boosting transferability.
Xie et al. \cite{xie2019improving} apply random transformations to the input images at each iteration, thus alleviating overfitting to the substitute model. Dong et al.\cite{dong2019evading} propose a translation-invariant attack method to generate more transferable adversarial examples against the defense models. Wang et al.\cite{Wang_2021_ICCV} admix input image with a small portion of each add-in image while using the original label of the input. Some works also concentrate on modifying the source model properly such as utilizing skip connections~\cite{wu2020skip} or the  back-propagation process~\cite{guo2020backpropagating} to boost transferability. Zhu et al. \cite{zhu2021rethinking} boost the transferability by matching the adversarial attacks with the directions which decrease the ground truth density.

As for targeted attacks,  Li et al.\cite{li2020towards} take adversarial examples away from the true label and push towards the target label, while \cite{inkawhich2019feature,Inkawhich2020Transferable} make the source image closer to an image of the target class in feature space, and then Inkawhich et al.\cite{inkawhich2020perturbing} generate targeted perturbations along with classifier information which transfers better than previous iterative methods. Gao et al.
 \cite{Cheng2021StatisticAlignment} align the source and target feature maps by high-order statistics with translation invariance. Zhao et al.
\cite{zhao2021success} use logit loss and generate perturbations by a large number of iterations. Usually, the iterative perturbation methods often take a long time to generate adversarial examples.

\emph{Generative Perturbations:} Another branch of attack uses generative models to craft adversaries. Compared with iterative methods, generative attacks achieve higher efficiency when attacking large-scale datasets, and they find adversarial patterns on a data distribution independent of a single sample. As long as the distribution of the clean images is fixed, the effectiveness of the adversarial examples depends on how the generator is trained. Therefore, many previous works about generative attacks aim to learn a strong generator, which maps the distribution of clean images to a distribution of adversarial examples that can generalize well on different black-box models. Our method follows this branch. GAP~\cite{poursaeed2018generative} proposes novel generative models to produce image-agnostic and image-dependent perturbations. CDA~\cite{naseer2019cross} uses relativistic training objective to boost cross-domain transferability. BIA~\cite{Zhang2022BIA} enhance the cross-domain transferability of adversarial examples from the data and model perspectives. TTP~\cite{naseer2021generating} achieves state-of-the-art performances in a targeted attack by maximizing the mutual agreement between the given source and the target distribution. 

However, to the best of our knowledge, the theoretical analysis of the transferability is still absent, while in this work we make the first attempt to provide a generalization error analysis for black-box targeted attack based on the model discrepancy. A new approach is also proposed based on our theoretical analysis, which achieves new state-of-the-art results for black-box targeted attack.

\noindent \textbf{Hypothesis Discrepancy:}
Ben-David et al.~\cite{NIPS2006_b1b0432c} define hypothesis discrepancy theory to measure the discrepancy of different models over a given distribution. Based on hypothesis discrepancy theory, the $\cH$-divergence~\cite{36364, NIPS2006_b1b0432c} and $\cH\Delta\cH$-divergence~\cite{36364, NIPS2006_b1b0432c} are proposed for measuring the difference between data distributions. These metrics are widely used in Domain Adaptation theory~\cite{36364, NIPS2006_b1b0432c, redko2019advances, acuna2021f}~ to derive upper bounds for generalization error. Many Domain Adaptation methods tried to minimize the generalization error upper bound by minimizing hypothesis discrepancy~\cite{saito2018maximum, lee2019sliced, Li21BCDM, lu2020stochastic}. These methods usually use multiple models to maximize and minimize the hypothesis discrepancy over the target domain such that the hypothesis discrepancy is small in the entire hypothesis set $\cH$. Different from these works, in this paper, we exploit the hypothesis discrepancy principle for analyzing black-box targeted attack.

\section{Methodology}
\label{sec:Methodology}

In this section, we first present the preliminary and  generalization error bound for black-box targeted attack in Section~\ref{sec:Preliminary} and \ref{sec:Generalization Error Bound}. Then we introduce the Minimizing Maximum Model Discrepancy (M3D) attack  in Section~\ref{sec:Theoretical Insight} and its implementation details in Section~\ref{sec:Overall Idea}.

\subsection{Black-Box Targeted Attack}
\label{sec:Preliminary}

In the black-box targeted attack task, we are given an unseen black-box classification model $D_b$ and an image dataset $\cA=\{(\x, y)\}$ with $\x$ being an image and $y$ being its class label. Normally, $D_b$ is trained to classify these images with  high accuracy, \ie, we expect $\arg\max_{c} D_b(\x) = y$. The goal is, given any image $\x$, to produce an adversarial example $\z$ to mislead the black-box model $D_b$ towards a specified class $y^t$ such that $\arg\max_{c} D_b(\z)= y^{t}$. Without loss of generality, the process of generating adversarial example can be denoted as $\z = G(\x)$, where $G$ can be implemented via an iterative optimization or a generator.

For ease of presentation, we define $\cX = \{\x\}$ as the set of images and the underlying distribution of images as $\cP$, namely, the image $\x \in \cX$ is deemed being sampled from the distribution $\cP$, \ie, $\x\sim\cP$. Correspondingly, the set of adversarial examples is denoted by $\cZ = \{\z\}$, and $\z\sim\cQ$ where $\cQ$ is the underlying distribution.

Since the black-box model $D_b$ is inaccessible, existing works usually construct adversarial examples based on a substitute model $D_s$. Owing to the transferability, these adversarial examples are often able to attack the black-box $D_b$ with a decent chance. Formally, constructing adversarial examples on a substitute model for black-box targeted attack can be written as,

\begin{equation}\label{eq:target_attack}
  \z = \mathop{\operatorname{argmin}}\limits_{\z} \ell\left(D_s\left(\z \right), y^{t}\right), \text { s.t. }\left\|\z-\x\right\|_{\infty} \leq \epsilon ,
\end{equation}
where the $l_\infty$-norm is applied to constrain the imperceptibility of the perturbation, and $\ell$ is the classification loss (\eg, cross-entropy loss). 

\subsection {Generalization Error Bound}
\label{sec:Generalization Error Bound}
Let us define $\mathcal{H}$ as a hypothesis set that contains all classifiers which perform well on the specified classification task. We also reasonably assume that the black-box model $h_b$ and an arbitrary substitute model $h_s$ satisfy $ h_b, h_s \in \cH$. Let us define $f_t$ as the labeling function for the misclassified target category, namely, $f_t$ classifies all images into the specified target category. 

The generalization error $\cE_{\cQ}(h_b, f_t) $ for the black-box targeted attack can be defined as:
\begin{equation}\label{eq: generalization error bound}
\begin{aligned} 
    \cE_{\cQ}(h_b, f_t) &= \underset{{\bf z}\sim \cQ}{\mathbb{E}} \left[ \ell{(h_b(\z), f_t(\z))}  \right], 
\end{aligned}
\end{equation}
where $\ell$ is the metric function that measures the prediction difference between $h_b$ and $f_t$ for a sample $\z$. Since we cannot access the black-box model $h_b$, we seek an upper bound of the generalization error $\cE_{\cQ}(h_b, f_t)$ as follows. 

\begin{thm}
For any $\delta\ge 0$, with probability $1-\delta$, we have the following generalization bound for the black-box classifier $h_b \in \mathcal{H}$ and any substitute classifier $h_s \in \cH$,
\begin{equation}\label{eq: optimize}
\begin{aligned}
\cE_{\cQ}(h_b, f_t) 
&\leq \hat{\cE}_{\cZ}(h_s, f_t) +  \sup _{h, h^{\prime} \in \cH}  \hat{\cE}_{\cZ}(h, h^{\prime})  +  \Omega,
\end{aligned}
\end{equation}
where $h$ and $h^{\prime}$ are two classifiers sampled from $\cH$, $\hat{\cE}$ is the empirical estimation of the generalization error, and $\Omega$ is a minor term. 
\end{thm}

The proof can be derived based on the $\cH$-divergence theory which was originally proposed for domain adaptation~\cite{NIPS2006_b1b0432c}. We sketch the proof here, and the completed proof is provided in the supplementary. By introducing the substitute model $h_s$, we have $\cE_{\cQ}(h_b, f_t) \leq \cE_{\cQ}(h_s, f_t) + \cE_{\cQ}(h_s, h_b)$. Moreover, from the definition of $\cH$-divergence, we have $\cE_{\cQ}(h_s, h_b) \leq \sup _{h, h^{\prime} \in \cH} \cE_{\cQ}(h, h^{\prime})$. Finally, we transform the generalization error $\cE_{\cQ}$ into an empirical estimation error $\hat{\cE}_{\cZ}$, where an additional minor term $\Omega$ is introduced which usually can be neglected in optimization. 

Despite the minor term $\Omega$, the above generalization bound mainly consists of two items: 1) the empirical error of attacking the substitute model; 2) the maximum empirical difference between any two models in the hypothesis set $\cH$. Most existing black-box targeted attack works assume the good transferability of the adversarial samples and mainly focus on designing techniques for attacking the substitute model. This can be seen to minimize the first term only in Eq.~(\ref{eq: optimize}) while ignoring the second term. However, the adversarial examples found in this way might not be robust enough to the variation of models, especially when there exists a model architecture difference between black-box and substitute models. Next, we will present a new approach for constructing robust adversarial samples by taking the second term into consideration as well.

\begin{figure*}
    \centering
    \includegraphics[width=0.85\linewidth]{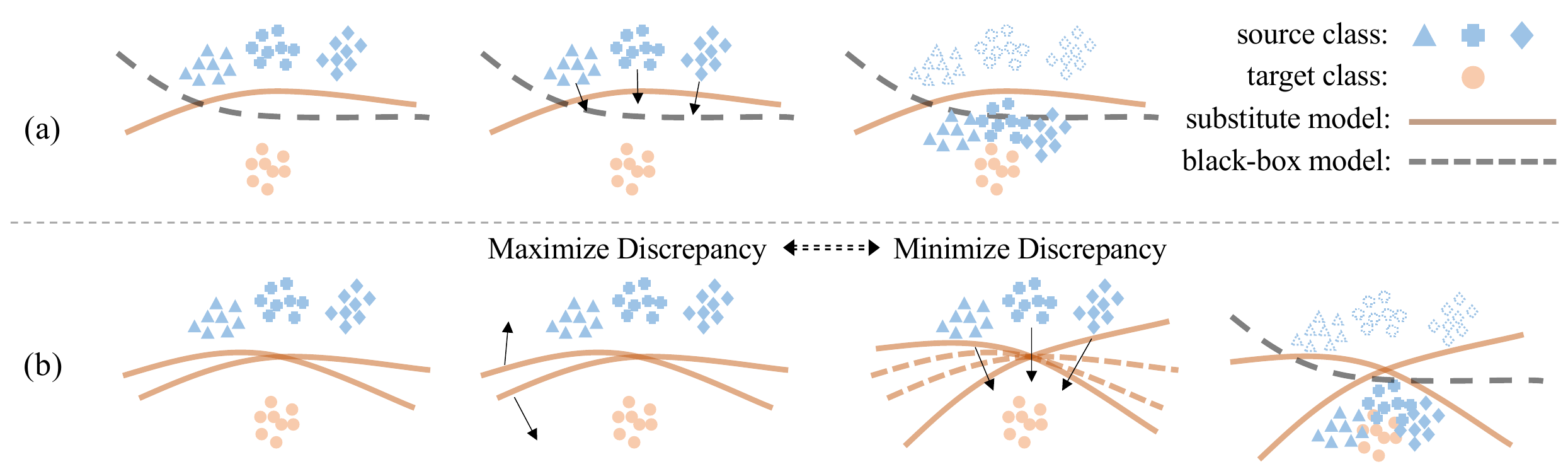}
    \caption{ Illustration on our M3D approach: (a) Traditional targeted attack methods craft adversarial examples~(moving blue plots towards orange plots) on one substitute model~(orange line), which might not be always effective due to the model discrepancy between the substitute model and the black-box model (black dash line); (b) Our M3D approach generates strong adversarial examples by a min-max game to attack two substitute models with maximum model discrepancy, which are more robust to the model discrepancy (Best viewed in color).}
    \label{fig:difference}
\vspace{-0.2cm}
\end{figure*}

\subsection{Minimizing Maximum Model Discrepancy}
\label{sec:Theoretical Insight}
As analyzed above, the robust adversarial examples for attacking a black-box model not only successfully attack the substitute model but also should preserve a minimum discrepancy between any two models in the hypothesis set.

To calculate the empirical maximum model discrepancy $\sup _{h, h^{\prime} \in \cH}  \hat{\cE}_{\cZ}(h, h^{\prime})$ (\ie, the second term in Eq.~(\ref{eq: optimize})), we introduce two discriminator models $D_1$ and $D_2$. Then, the empirical maximum model discrepancy can be calculated by
\begin{equation}
  \scalebox{0.95}{$\displaystyle \max_{D_1,D_2}\underset{{\bf x}\in \cX}{ \hat{\mathbb{E}} } d\left[D_{1}\circ G({\bf x}) , D_{2}\circ G({\bf x}) \right]$}\label{eq:Discriminator},
\end{equation}
where $\hat{\mathbb{E}}$ represents the empirical estimator, \ie, $\hat{\mathbb{E}}_{\x\in \cX} d(\cdot, \cdot)= \frac{1}{|\cX|}\sum_{\x\in \cX}d(\cdot, \cdot)$, and  $d(\cdot, \cdot)$ is the discrepancy metric between the $D_1$ and $D_2$. 
We use the $\ell_1$ distance as the discrepancy metric. Note that here we rewrite the empirical estimator $\hat{\cE}_{\cZ}$ as $\hat{\mathbb{E}}_{\x\in \cX}$, as each adversarial example $\z \in \cQ$ is generated correspondingly from the clean image $\x \in \cX$. 

On the other hand, the robust adversarial examples are expected to preserve a minimum discrepancy for any two models, which leads to the following min-max objective,
\begin{equation}
  \scalebox{0.95}{$\displaystyle \mymin_{G} \mymax_{D_1,D_2} \cL_d = \underset{{\bf x}\in \cX}{\mathbb{\hat{E}}} d\left[D_{1}\circ G({\bf x}) , D_{2}\circ G({\bf x})\right] $}\label{eq:min-max}.
\end{equation}

Note the adversarial examples are also expected to successfully attack a given substitute $h_s\in \cH$. As $h_s$ can be any classifier, without loss of generality, we use both $D_1$ and $D_2$ as the substitute models. Therefore, the generator should be trained to mislead the two discriminators to classify the adversarial example into the specified category $y^t$, 
\begin{equation}
\!\!\!\!\!\!\min_{G}\cL_{a} \!\!=\!\! \underset{{\x}\in \cX}{\mathbb{\hat{E}}}   \ell_{ce}\left( D_{1} \circ G(\x), {y^t}\right) + \ell_{ce}\left( D_{2} \circ G(\x), {y^t}\right)\!,\!
\label{eq:attack_loss}
\end{equation}
where $\ell_{ce}(\cdot, \cdot)$ is the cross-entropy loss. 

\begin{figure*}
    \centering
    \includegraphics[width=0.88\linewidth]{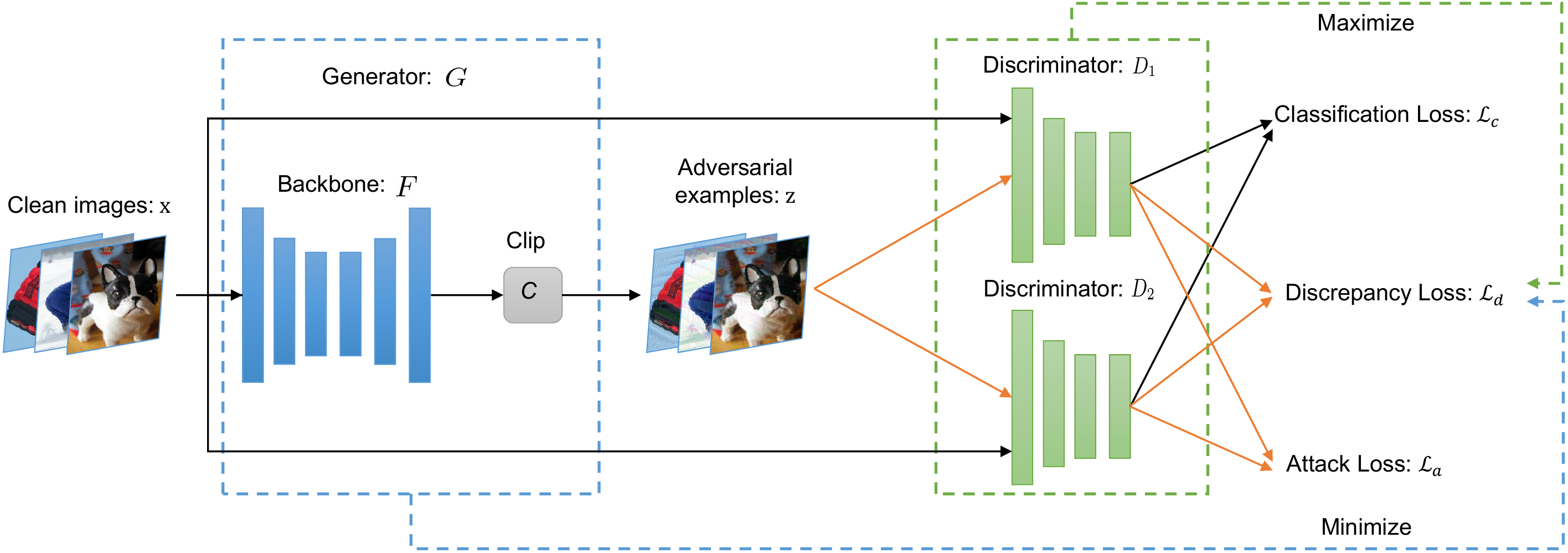}
    \caption{Overview of our M3D approach for black-box targetd attack, which consists of a generator and two discriminators. The discriminators act as the substitute models and are trained to keep a large discrepancy on the adversarial examples. The generator is trained to generate robust adversarial examples that are able to attack two discriminators and keep their discrepancy as small as possible. The gray rectangle labeled with $C$ is the clipping operation. See the description in Section~\ref{sec:Overall Idea} for details. 
    }
    \label{fig:framework}
\end{figure*}

Moreover, as we presume that the hypothesis set $\cH$ consisting of classifiers that perform well on the specified classification task, these two discriminators $D_1$ and $D_2$ should maintain a low classification loss on the clean data as well, 
\begin{equation}\label{eq: classification loss}
\min _{D_1, D_2} \mathcal{L}_{c} = \underset{{(\x,y)}\in \cA}{\mathbb{\hat{E}}} 
\left [ \ell_{ce}{(D_1(\x), y)} +  \ell_{ce}{(D_2(\x), y)} \right],
\end{equation}
where $\ell_{ce}(\cdot, \cdot)$ is the cross-entropy loss. By taking the losses in Eq.~(\ref{eq:min-max}), Eq.~(\ref{eq:attack_loss}) and Eq.~(\ref{eq: classification loss}) into consideration, the overall objective function of our minimizing maximum model discrepancy (M3D) attack approach can be written as,
\begin{equation}\label{eq: overall loss}
 \min _{G} \cL_{a} + \max_{D_1, D_2} [\cL_d - \cL_{c}].
\end{equation}

By optimizing the above objective function, we can learn a strong generator $G(\x)$, which maps the clean images to the adversarial examples that are robust to the model variation. We illustrate the intuition behind our M3D approach in  Fig.~\ref{fig:difference}. Previous targeted attack works mainly concentrate on crafting adversarial examples on a fixed substitute model~(orange line), which can be seen as moving the samples from source classes~(blue plots) towards the samples from the target class~(orange plots) by adding small perturbations on the original images. However, these adversarial examples may fail to attack the black-box model~(black line) due to the model discrepancy between the substitute and the black-box model. In contrast, in our M3D approach, we generate strong adversarial examples to attack two substitute models, which are trained dynamically to maintain a maximum discrepancy for predicting these adversarial examples. In this way, the adversarial examples would move closer to the target class thus being more robust to the variation of the black-box model.
\begin{algorithm}
\small
\caption{Minimizing Maximum Model Discrepancy for Black-box Targeted Attack}
\label{alg: Algorithm}
\begin{algorithmic}[1]
\Require Training data $\cA$, pretrained substitute models $D_1$,$D_2$, perturbation budget $\epsilon$. 
\State Randomly initialize the generator $G$

\Repeat

\State Randomly sample a mini-batch of training data $\tilde{\cA} = \{(\x, y) \} \subset \mathcal{A}$, and denote mini-batch of images as $\tilde{\cX} = \{x\}$.

\State Forward-pass all images $\x \in \tilde{\cX} $  through the generator and generate bounded adversaries $\{\z\}$ using Eq.~(\ref{eq: smooth projection}). 

\State Forward pass all $\z$ through $D_1$,$D_2$. 
\State \textbf{Update $G$:}
\State Calculate the loss $\cL_a + \cL_d$ and perform backward pass to update $G$ while keeping $D_1$,$D_2$ frozen. 
\State \textbf{Update $D_1$ and $D_2$:}

\State Forward pass all clean images $\x$'s  through $D_1$,$D_2$. 
\State Calculate the loss $\cL_c - \cL_d$, and perform backward pass to update $D_1, D_2$, while keeping $G$ fixed. 

\Until{ The stop criterion is reached.}
\end{algorithmic}
\end{algorithm}

\subsection{Implementation}
\label{sec:Overall Idea}

We depict the training pipeline of our M3D approach in Fig.~\ref{fig:framework}, which consists of a generative model $G$ and two discriminators $D_1$ and $D_2$. The clean images pass through $G$ to get their adversarial versions, and then these adversarial examples are respectively fed into $D_1$ and $D_2$.
During training, we alternatively train the discriminators and the generator for an amount of iterations. For the discriminators, we fix $G$ and train them to minimize the classification loss on the clean images and simultaneously maximize the discrepancy between the two discriminators. For the generator, we fix $D_1$ and $D_2$ and train $G$ to attack both $D_1$ and $D_2$ and at the same time minimize the discrepancy between the two discriminators. The whole training procedure is shown in Algorithm \ref{alg: Algorithm}. We detail the design of the generator and discriminators as follows. 

\textbf{Generator:}
For the generator, we use the same network architecture as in previous generative attack works~\cite{poursaeed2018generative, naseer2021generating, naseer2019cross}.
The generator consists of a backbone $F$ and a clip operation $C$.
The generator backbone consists of downsampling, residual and upsampling blocks whose output is an adversarial sample with the same size as of input. The detail of the generator network is provided in the supplementary. As the adversarial noise $\boldsymbol{\delta}=|| G(x) - x ||$ is strictly constrained under a norm distance for the imperceptibility of
 perturbation \ie, $\|\boldsymbol{\delta}\|_{\infty} \leq \epsilon$, we follow~\cite{naseer2021generating} to scale the output of $F$ by using a differentiable clipping operation. 
\begin{equation}\label{eq: smooth projection}
    \boldsymbol{\z}=\operatorname{clip}\left(\min \left(\boldsymbol{\x}+\epsilon, \max \left(\mathcal{W} * F\left(\boldsymbol{\x}\right), \boldsymbol{\x}-\epsilon\right)\right)\right),
\end{equation}
where $\mathcal{W}$ is a smoothing operator with fixed weights that reduces high frequencies without violating the $l_\infty$ distance constraint.

\textbf{Discriminators:} The discriminators can be any classification model. We follow the previous targeted attack works~\cite{naseer2021generating} to attack a black-box model trained on ImageNet, so we initialize the discriminators with the ImageNet pretrained classification models using different architectures such as ResNet~\cite{he2016deep}, DenseNet~\cite{huang2017densely}, VGGNet~\cite{simonyan2015very} for an extensive evaluation (see Section~\ref{sec:results} for details).

\input{tables/comparison_on_generative.tex}

\section{Experiments}

\subsection{Experiments Setup}\label{sec:setup}

For a fair comparison, we follow the recent state-of-the-art method TTP~\cite{naseer2021generating} to evaluate our method. We adopt the same architecture for the generator $G$ and vary the substitute model with different architecture as similar as in TTP~\cite{naseer2021generating}. Note that in each round of experiments (\eg, using ResNet50 as the substitute model), the substitute model in all methods shares the same network architecture to ensure a fair comparison. Moreover, the two discriminator models $D_1$ and $D_2$ in our M3D approach also share the same network architecture.
For the training data, we use 49.95K images sampled from ImageNet train set~\cite{russakovsky2015imagenet}. For the testing data, we adopt the original ImageNet validation set (50k samples).  We use Adam optimizer~\cite{kingma2014adam} with a learning rate of 0.0002 and the values of exponential decay rate for the first and second moments are set to 0.5 and 0.999, respectively. We conduct all the experiments on GeForce RTX 3090 GPUs with a PyTorch~\cite{pytorch_cite} implementation.

We evaluate our approach on both subset-source setting~\cite{Inkawhich2020Transferable} and all-source setting~\cite{naseer2021generating}. In the subset-source setting, 10 classes in the test data are used for evaluation. Each round, we select one of these 10 classes as the target class~($50$ samples), and the remaining 9 classes~($450$ samples) are the samples that need to be attacked. In the all-source setting, 1000 classes in the test data are used for evaluation. Each round, we select one of these 1000 classes as the target~($50$ samples), and the remaining 999 classes~($49{,}950$ samples) are the samples that need to be attacked.
For both settings, we evaluate our model at each round and report the average Top-1(\%) target accuracy over all rounds, where the Top-1(\%) target accuracy means the proportion of adversarial examples that the black-box model predicts to the target class. 

\begin{figure*}

    \centering
    \includegraphics[width=0.8\linewidth]{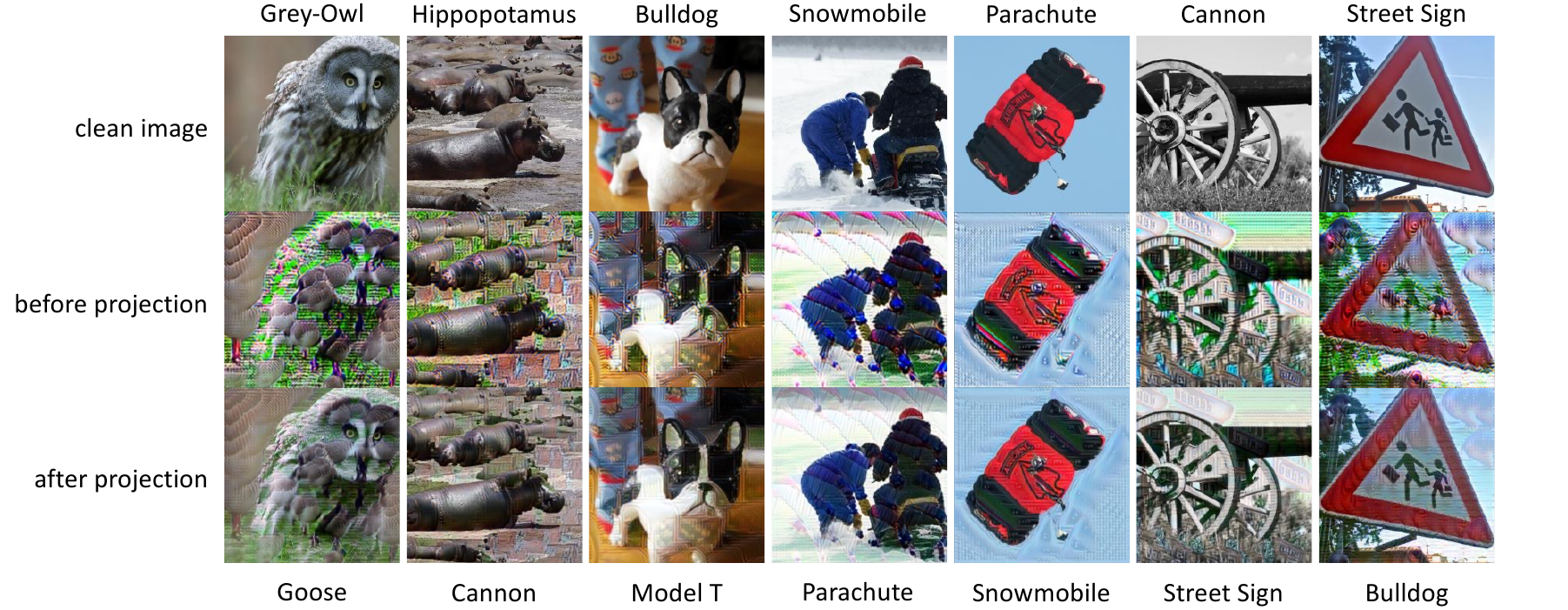}
    \caption{
    Targeted adversaries produced by generator trained against ResNet50. 1st row shows original images while 2nd row shows unrestricted outputs of an adversarial generator, and 3rd row shows adversaries after valid projection. The caption in the top represents the original category. The caption below represents the target category. Perturbation budget is set to $l_\infty \le 16$.
    }
    \label{fig:visualization}
\vspace{-0.5cm}
\end{figure*}

\subsection{Comparison with State-of-the-art Methods}\label{sec:results}

\subsubsection{Attacking Unknown Models}
We first evaluate our proposed approach for attacking unseen classification models pretrained on ImageNet with both subset-source and all-source settings. Following~\cite{naseer2021generating}, we respectively conduct experiments using the VGG19$_{BN}$, DenseNet121, ResNet50 to initialize our substitute models, and attack VGG19$_{BN}$, DenseNet121, and ResNet50 models. The ResNet152 and WRN-50-2 are additionally used as the black-box models in the all-source setting. We follow~\cite{naseer2021generating} to compare with a bunch of baselines including  MIM~\cite{dong2018boosting}, AA~\cite{inkawhich2019feature}, FDA~\cite{Inkawhich2020Transferable}, DIM~\cite{xie2019improving}, GAP~\cite{poursaeed2018generative}, CDA~\cite{naseer2019cross} and TTP~\cite{naseer2021generating}. 

The visualization of our adversarial examples is shown in Fig.~\ref{fig:visualization}.
The results of all methods on different settings are shown in Table~\ref{tab: undefended_models_0511}, where the results of baselines are from \cite{naseer2021generating}. We observe that our proposed approach outperforms all existing methods for black-box targeted attack with a clear margin. 
Specifically, in the all-source setting, our method outperforms the recent state-of-the-art TTP~\cite{naseer2021generating} for black-box targeted attack by $12.75\%$ to 41.23\%. 
When the substitute models are VGG19$_{BN}$, DenseNet121, ResNet50, our method averagely outperforms TTP~\cite{naseer2021generating} by $36.69\%$, $31.56\%$, $13.98\%$ respectively.
It is worth noting that our method shows impressive improvements when the black-box model has a large model discrepancy from the substitute model. 

For example, when the substitute model is VGG19$_{BN}$ and the black-box model is ResNet152, our method can reach $68.41\%$ target accuracy while the TTP method is $27.18\%$.
A similar tendency can also be observed in the subset-source setting.
These results clearly demonstrate the effectiveness of our methods for black-box targeted attack. 

\input{tables/comparison_on_defended.tex}

\subsubsection{Attacking Unknown Robust Models}

To further evaluate the robustness of the adversarial examples, we follow TTP~\cite{naseer2021generating} to attack unknown robust models. Three kinds of robust models are tested under the all-source setting: 1) models trained with data augmentation methods to be robust to natural corruptions, such as Augmix~\cite{hendrycks2020augmix}; 2) models trained on stylized ImageNet to have a higher shape bias so that they are inherently more robust to image style distortions, such as SIN and SIN-IN~\cite{geirhos2018imagenettrained}; 3) adversarially trained models which are shown to be able to defend against adversarial examples to certain extent~\cite{madry2018towards,salman2020adversarially}. We craft adversarial examples by using the ResNet50 architecture for the substitute models and then attack the ResNet50 models trained by different robust training mechanisms.

The results of all methods are shown in Table~\ref{tab: augmentation_vs_stylized_vs_adv}, where the numbers for baseline methods are from \cite{naseer2021generating}. We observe that our results outperform previous generative attacks in all settings. Particularly, we can see a significant improvement of 17.0\%, 35.19\%, 21.73\% when attacking Augmix~\cite{hendrycks2020augmix} trained model, stylized ImageNet~\cite{geirhos2018imagenettrained} models, adversarially~\cite{madry2018towards,salman2020adversarially} trained models, respectively. 
These results prove that our adversarial examples are robust not only to the variation of model architecture but also to popular defense mechanisms.

\subsubsection{Attacking Real-world System}
\label{sec: Google Cloud Vision}
We also evaluate our methods in a more challenging task, \ie,  to fool a real-world computer vision system Google Cloud Vision. 
As for each image, the Google Cloud Vision API will return a list of the ten most likely classes. 
For targeted attack, as long as the target class is in the returned list, we consider it a success. As the semantic labels predicted by the API don't correspond to the 1k ImageNet labels, we consider semantically consistent classes as the same class~(The criteria of \emph{``semantically consistent''} is shown in Table~\ref{tab: G_cloud} with our results).
As for evaluation, we randomly sample 100 adversarial examples crafted by TTP~\cite{naseer2021generating} and our M3D method, respectively. 
Specifically, the adversarial examples are generated on ResNet50. 
From Table~\ref{tab: G_cloud}, we observe that even the real-world vision system Google Cloud Vision can be attacked successfully by our approach with a certain chance. This reveals the vulnerability of real-world systems.
Our M3D approach outperforms TTP~\cite{naseer2021generating} on all classes with a significant difference. 
\begin{table}[ht]
\vspace{0.3cm}
\centering
\caption{ Target Transferability on Google Cloud Vision:   Top-1(\%) target accuracy averaged across 100 images. Perturbation budget: $l_{\infty}\leq 16$ .}
\resizebox{80mm}{12mm}{
\begin{tabular}{cccc}
\toprule
Target Class & Similar Class        & TTP & Ours \\ \hline
Grey-Owl     & Owl, Screech owl, great grey owl              & 0   & 24   \\ 
Goose        & Goose, geese    & 4   & 37   \\ 
Bulldog      & Dog, Companion dog       & 25  & 61   \\ 
Parachute    & Parachute            & 4   & 23   \\ 
Street sign  & Signage, Street sign & 26   & 47  \\ \bottomrule
\end{tabular}
}

\label{tab: G_cloud}
\end{table}

\begin{table}[t]
    \centering
\small
    \caption{Target Transferability of Perturbations: Top-1 (\%) target accuracy in the subset-source setting.}\label{tab:perturbation_attack}
\resizebox{72mm}{15mm}{
      \begin{tabular} {ccccc}
				\toprule

			Src.&Attack&VGG19$_{BN}$ & Dense121 & ResNet50\\
			\midrule
			VGG19$_{BN}$ & \makecell[l]{TTP ~\cite{naseer2021generating} \\ Ours}& \makecell[c]{75.11 \\\textbf{78.78}} &\makecell[c]{9.36 \\\textbf{28.18}}& \makecell[c]{10.58\\\textbf{17.62}} \\
			\midrule
			
				Dense121 & \makecell[l]{TTP ~\cite{naseer2021generating} \\ Ours}& \makecell[c]{19.44 \\\textbf{42.53}} &\makecell[c]{58.58 \\\textbf{69.98}}& \makecell[c]{16.20\\\textbf{31.56}} \\
			\midrule
			
				ResNet50 & \makecell[l]{TTP ~\cite{naseer2021generating} \\ Ours}& \makecell[c]{25.02 \\\textbf{45.62}} &\makecell[c]{33.96 \\\textbf{53.73}}& \makecell[c]{40.69\\\textbf{54.56}} \\
			\bottomrule
	\end{tabular}
}
\end{table}

\begin{figure}[t]
\centering
   \includegraphics[width=0.9\linewidth]{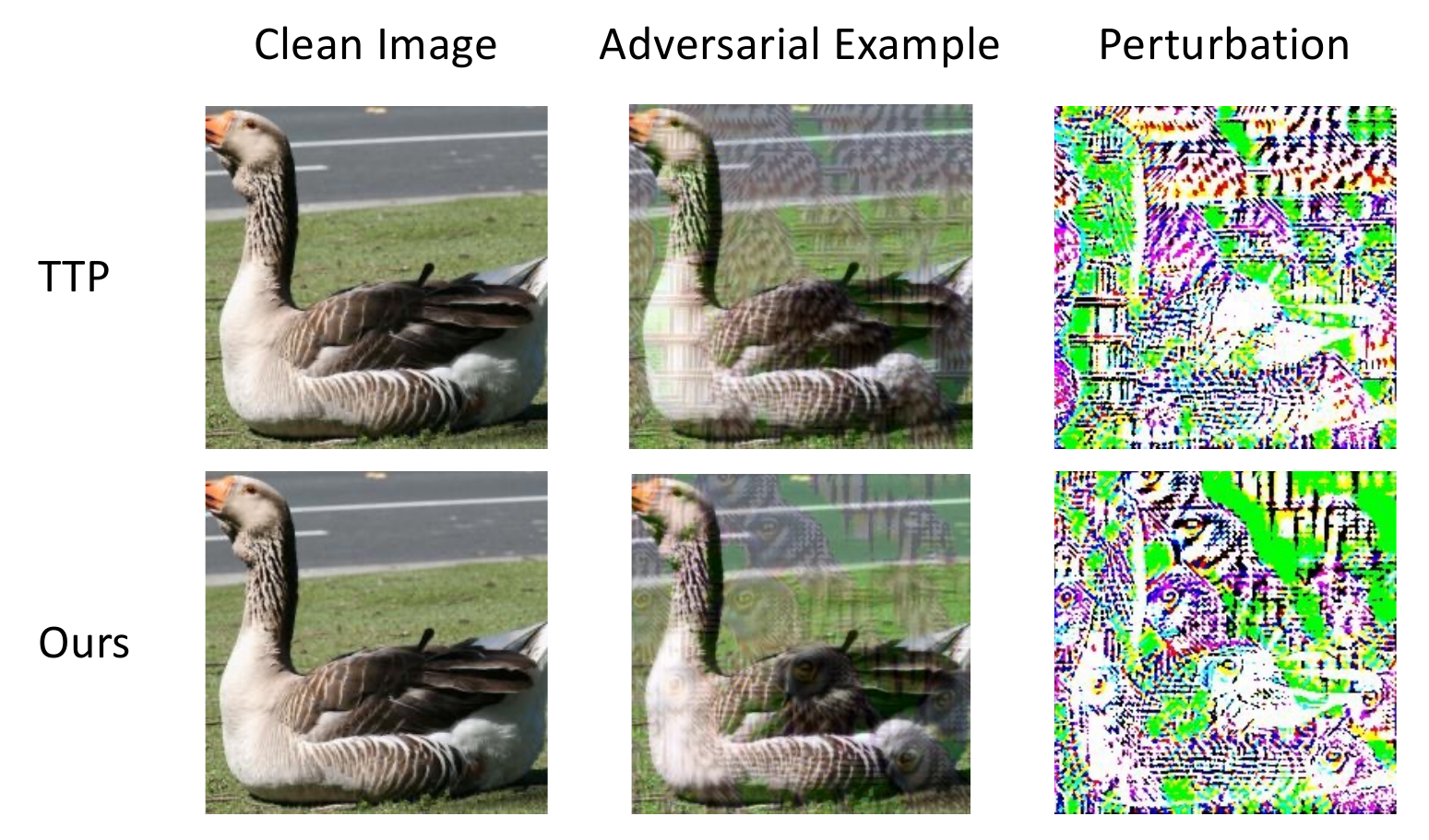}

\caption{The perturbation comparison with TTP~\cite{naseer2021generating} 1st column shows the clean images while 2st column shows the clipped adversarial examples, and 3rd column shows the perturbations. }
\label{fig:compare_ttp}
\vspace{-0.4cm}
\end{figure}

\subsection{Additional Analysis}\label{sec:Analysis}

\textbf{Perturbation Robustness:} We experimentally find that the generated adversarial semantic pattern itself generalizes well among the different models. We illustrate and corroborate our claim by directly feeding scaled adversarial perturbations into different classifiers in the subset-source setting. The result in Fig.~\ref{fig:compare_ttp} and Table~\ref{tab:perturbation_attack} reveal that the perturbations generated by our method can be considered as the target class and have a higher attack success rate compared with those by TTP~\cite{naseer2021generating}.

 \textbf{Perturbation Effectiveness:} In order to understand the improvements of our method, we put a further analysis by taking the classification error rate into consideration.

 The classification error rate is defined as the proportion of adversarial examples that are not classified as their original classes by the black-box model, which can be used to measure the ability to move samples away from their original classes. However, this does not necessarily mean the targeted attack is successful since we expect the black-box model could classify these samples to the specified class.

\begin{figure}

    \centering
    \includegraphics[width=0.9  \linewidth]{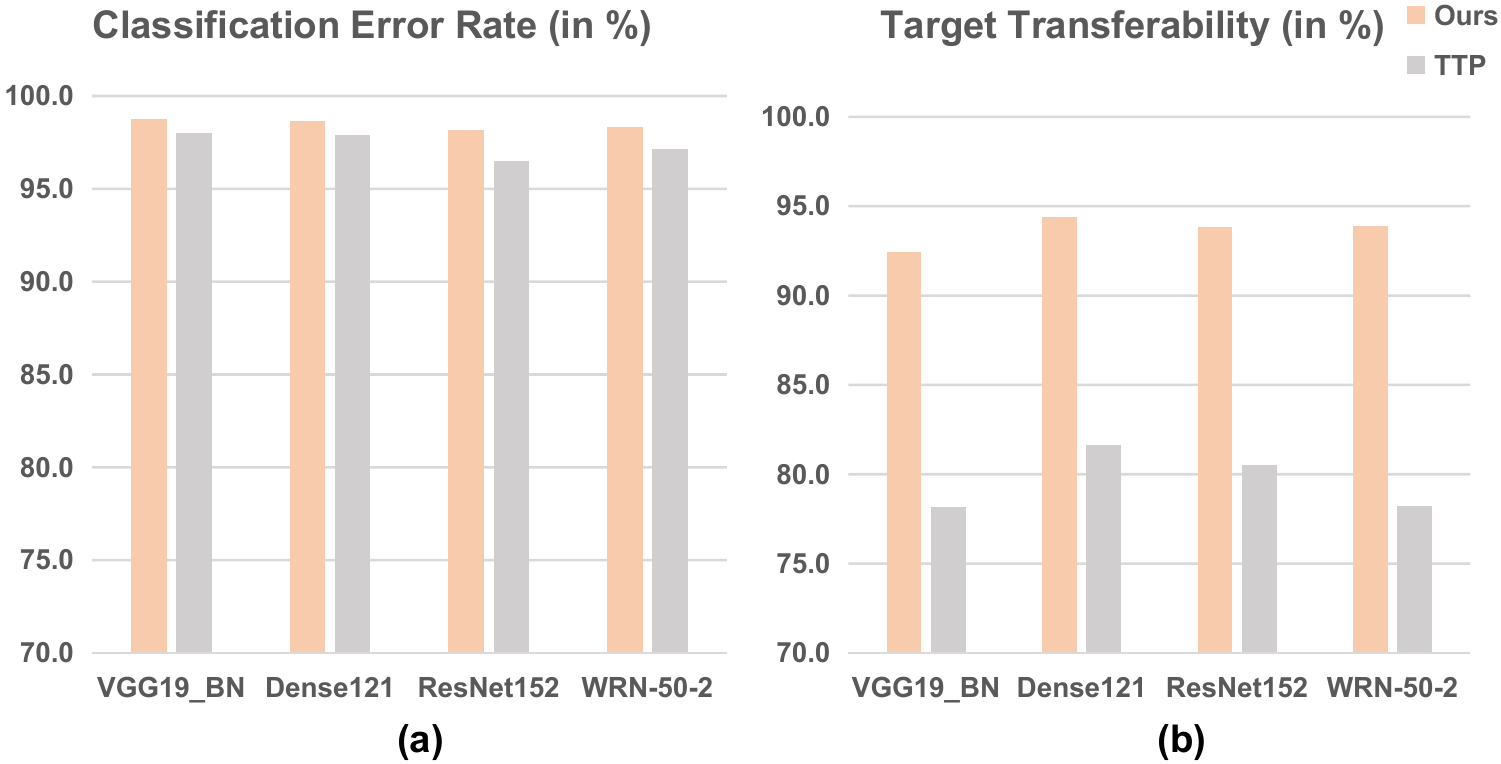}
    \caption{ Classification Error Rate and Target Transferability in the all-source setting under the perturbation budget  $l_{\infty}\leq 16$. Though having similar classification error rate, our method outperforms TTP in target transferability.
    }
    \label{fig:histogram}
\end{figure}

 It can be seen from Fig.~\ref{fig:histogram} that, though both TTP~\cite{naseer2021generating} and our method have the similar ability to mislead the target black-box model to predict wrong categories, our method outperforms TTP~\cite{naseer2021generating} by a significant margin at attack success.
 This confirms our analysis in the last paragraph of Section~\ref{sec:Theoretical Insight} that our M3D approach help to draw the adversarial examples closer to the target class.

\subsection{Ablation Studies} 
To gain a better understanding of the effect of our M3D approach, we present an ablative study in Table.~\ref{tab:ablation}. We consider three cases: A) using one fixed substitute model; B) using two trainable models without considering their model discrepancy; C) the full model of our M3D method. For all cases, the substitute models are initialized using the ResNet50 pretrained on ImageNet, and we attack four normal models and one robust model. As shown in Table.~\ref{tab:ablation}, compared to using one fixed substitute model, using two trainable models gains improvements due to the model ensemble, however, is still inferior to the full model with a clear margin. This clearly demonstrates the effectiveness of minimizing the maximum model discrepancy (\ie, the second term in Eq.~(\ref{eq: optimize})) in our M3D approach. 

\begin{table}
\vspace{-0.2cm}
\caption{Ablation study in the all-source setting. 'A' indicates training the generator with one fixed substitute model. 
'B' indicates training the generator with two trainable substitute models without considering model discrepancy. 'C' indicates training the generator with our method M3D. The substitute model is chosen as ResNet50.}
\begin{center}
\small
  \scalebox{0.9}{

\begin{tabular}{cccc}
\toprule
             & A     & B                         & C                         \\ \midrule
Number       & 1     & 2                         & 2                         \\ 
Trainable    & -     & \checkmark & \checkmark \\ 
Discrepancy  & -     & -                         & \checkmark \\ \midrule
VGG19$_{BN}$ & 71.26 & 76.86                     & \bf{92.41}                     \\ 
Dense121     & 81.73 & 85.07                     & \bf{94.39}                     \\ 
ResNet152    & 78.36 & 79.51                     & \bf{93.85}                     \\ 
WRN-50-2     & 78.88 & 80.19                     & \bf{93.87}                     \\ 
SIN          & 24.18 & 31.48                     & \bf{65.36}                     \\ \bottomrule
\end{tabular}
}
\end{center}
\label{tab:ablation}
\vspace{-0.8cm}
\end{table}

\section{Conclusion}

In this paper, we study the black-box targeted attack problem from the model discrepancy perspective. We first present a generalization error bound based on the model discrepancy for the black-box targeted attack. 
Based on this bound, we design a novel approach called Minimizing Maximum Model Discrepancy~(M3D) attack, which is capable of crafting highly transferable adversarial examples by minimizing the maximum model discrepancy between two substitute models. We show that our M3D approach can be easily implemented with deep neural networks, which are trained in the adversarial training manner between the generator and the substitute models. Extensive experiments demonstrate that our proposed M3D approach outperforms existing state-of-the-art methods with a significant margin for the black-box targeted attack using various models on the ImageNet dataset.

{\small
\bibliographystyle{ieee_fullname}
\bibliography{egbib}
}

\end{document}

%% file: tables/comparison_on_generative.tex
\begin{table*}[!htp]
        \vspace{-0.2cm}
    \caption{ Target Transferability Against Unkown Models:  Top-1(\%) target accuracy in all-source and sub-source setting . Perturbation budget: $l_{\infty}\leq 16$ . In comparison to previous state-of-the-art methods, our method outperforms them by a large margin.  '$*$' indicates white-box attack. In subset-source setting, only black-box attack results are shown.   }
    \setlength{\tabcolsep}{4pt}
    \centering
    \scalebox{0.72}{
    \begin{tabular}{llccccc|lccc}
    \toprule
    &\multirow{2}{*}{\textbf{{Attack}}}   & \multicolumn{9}{c}{\textbf{Naturally Trained}}\\
    \cmidrule(lr){3-11} 
    &    & \multicolumn{5}{c}{\textbf{All-source}}  &  \multicolumn{4}{c}{\textbf{Subset-source}}  \\
          \cmidrule(lr){2-7}  \cmidrule(lr){8-11}
          & \textbf{{Method}} & \textbf{VGG19$_{BN}$} & \textbf{Dense121} & \textbf{ResNet50} & \textbf{ResNet152} & \textbf{WRN-50-2} & \textbf{{Method}} &  \textbf{VGG19$_{BN}$}  &  \textbf{Dense121} &  \textbf{ResNet50}\\
         \midrule
       \parbox[t]{2mm}{\multirow{6}{*}{\rotatebox[origin=c]{90}{VGG19$_{BN}$}}}

        & MIM~\cite{dong2018boosting}&\textbf{99.91}$^*$&0.92&0.68&0.36&0.47 & AA~\cite{inkawhich2019feature} & --&0.8&0.6 \\
        & DIM~\cite{xie2019improving}&99.38$^*$&3.10&2.08&1.02&1.29   & FDA-fd~\cite{Inkawhich2020Transferable} &-- & 3.0&2.1  \\
         & GAP~\cite{poursaeed2018generative}  & 98.23$^*$& 16.19& 15.83&  5.89& 7.78  & FDA$^N$~\cite{inkawhich2020perturbing} & -- & 6.0 & 5.4  \\
          & CDA~\cite{naseer2019cross} & 98.30$^*$& 16.26& 16.22&  5.73&  8.35 & CDA~\cite{naseer2019cross}&--&17.82&17.09 \\
        & TTP~\cite{naseer2021generating} & 98.54 $^*$ & 45.77& 45.87  & 27.18
            & 32.63 & TTP~\cite{naseer2021generating}& --&48.29& 47.07\\
         \cmidrule(lr){2-11}

          & Ours & 99.22$^*$ & \textbf{79.46} & \textbf{81.91}  &\textbf{68.41} & \textbf{68.43} &Ours&\textbf{--}&\textbf{80.47}& \textbf{81.44} \\

         \midrule
        
         \parbox[t]{2mm}{\multirow{6}{*}{\rotatebox[origin=c]{90}{Dense121}}} 
        & MIM~\cite{dong2018boosting}&1.85&\textbf{99.90}$^*$&2.71&1.68&1.88 & AA~\cite{inkawhich2019feature} & 0.0 & -- & 0.0\\
			& DIM~\cite{xie2019improving}&7.31&98.81$^*$&9.06&5.78&6.29 & FDA-fd~\cite{Inkawhich2020Transferable}&34.0 & --& 34.0\\
         & GAP~\cite{poursaeed2018generative}  & 39.01&  97.30$^*$& 47.85& 39.25&  34.79 & FDA$^N$~\cite{inkawhich2020perturbing}  & 42.0 & -- & 48.3 \\
          & CDA~\cite{naseer2019cross} &42.77&  97.22$^*$& 54.28 &   44.11 &  46.01 & CDA~\cite{naseer2019cross}&44.84&--&53.73 \\
        
        & TTP~\cite{naseer2021generating} & 58.90& 97.61$^*$&68.72  & 57.11 & 56.80 	& TTP~\cite{naseer2021generating}&61.75&--&69.60\\
          \cmidrule(lr){2-11}
        
        & Ours & \textbf{92.73} & 98.60$^*$ & \textbf{94.23}  &\textbf{90.06} & \textbf{90.75} & Ours&\textbf{93.02}&--&\textbf{95.22} \\
         
         \midrule
          \parbox[t]{2mm}{\multirow{6}{*}{\rotatebox[origin=c]{90}{ResNet50}}} 
          & MIM~\cite{dong2018boosting}&1.58&3.37&98.76$^*$&3.39&3.17 & AA~\cite{inkawhich2019feature} & 1.1 & 2.0 & --\\
			& DIM~\cite{xie2019improving}&9.14&15.47&\textbf{99.01}$^*$&12.45&12.61 & FDA-fd~\cite{Inkawhich2020Transferable}&16.0&21.0&--\\
          & GAP~\cite{poursaeed2018generative}  & 58.47& 71.72&  96.81$^*$& 64.89&  61.82 & FDA$^N$~\cite{inkawhich2020perturbing} &32.1&48.3&--\\
          & CDA~\cite{naseer2019cross} & 64.58&  73.57&  96.30$^*$& 70.30& 69.27 	& CDA~\cite{naseer2019cross} & 68.55&75.68&--\\
        
        & TTP~\cite{naseer2021generating} & 78.15& 81.64& 97.02$^*$  & 80.56
            & 78.25  & TTP~\cite{naseer2021generating}& 79.04&84.42&--\\
          \cmidrule(lr){2-11}

         & Ours & \textbf{92.41}& \textbf{94.39}& 98.33$^*$& \textbf{93.85}&\textbf{93.87} & Ours&\textbf{92.24}&\textbf{95.36}&--\\
         \midrule
    \end{tabular}}

    \label{tab: undefended_models_0511}
\end{table*}

%% file: tables/comparison_on_defended.tex
\begin{table}
    \caption{Target Transferability Against Unknown Robust Models:
  Top-1 (\%) target accuracy in all-source setting. Generators are trained against naturally trained ResNet50 . Perturbation are then transferred to ResNet50 trained using different methods including Augmix \cite{hendrycks2020augmix},  Stylized  \cite{geirhos2018imagenettrained} or adversarial \cite{salman2020adversarially}. 
  }
  
  \centering 
  \scalebox{0.67}{
  \begin{tabular}{llccccccc} 
    \toprule
     
     \multirow{3}{*}{$\epsilon$} & \multirow{3}{*}{Attack}    & \multirow{3}{*}{Augmix~\cite{hendrycks2020augmix}}   &  \multicolumn{2}{c}{Stylized \cite{geirhos2018imagenettrained}}   & \multicolumn{4}{c}{Adversarial \cite{salman2020adversarially}}  \\
     \cmidrule(lr){4-5}  \cmidrule(lr){6-9}
      &  &  & \multirow{2}{*}{SIN-IN} &  \multirow{2}{*}{SIN}  &   \multicolumn{2}{c}{$l_\infty$ }&\multicolumn{2}{c}{$l_2$} \\
       \cmidrule(lr){6-7} \cmidrule(lr){8-9}
      && &       &  & $\epsilon{=}.5$&  $\epsilon{=}1$&$\epsilon{=}.1$&$\epsilon{=}.5$ \\   
    \midrule
    \multirow{4}{*}{16}  & GAP \cite{poursaeed2018generative} &51.57&76.92&12.96&1.88&0.34&23.41&0.92 \\
    &CDA \cite{naseer2019cross} & 59.79& 75.93&9.21&2.10&0.39&23.89&1.18\\
     &TTP \cite{naseer2021generating}& 73.09&87.40&30.17&4.63&0.56&45.40&1.99\\
    \cmidrule(lr){2-9}

     &Ours  &\textbf{90.09}&\textbf{95.57}&\textbf{65.36}&\textbf{7.82}&\textbf{0.85}&\textbf{67.13}&\textbf{4.10}\\
     
    \midrule
    \multirow{4}{*}{32}  & GAP \cite{poursaeed2018generative} & 54.86 & 81.15&28.07&26.32&6.36&59.04&16.53\\
    &CDA \cite{naseer2019cross} &63.18&76.81&19.65&27.60&6.74&57.54&16.07 \\
     &TTP \cite{naseer2021generating}& 78.66&91.27&41.52&46.82&16.35&75.97&30.94\\
    \cmidrule(lr){2-9}

     &Ours  &\textbf{94.26} &\textbf{97.53} &\textbf{81.71} &\textbf{65.20}&\textbf{22.84}&\textbf{92.19}&\textbf{44.48}\\
     
    \bottomrule
  \end{tabular}}
  \label{tab: augmentation_vs_stylized_vs_adv}
  \vspace{-0.5cm}
\end{table}